\documentclass{article}

\PassOptionsToPackage{numbers, compress}{natbib}
\usepackage[preprint]{neurips_2026}

\usepackage[utf8]{inputenc}
\usepackage[T1]{fontenc}
\usepackage{hyperref}
\usepackage{url}
\usepackage{booktabs}
\usepackage{amsfonts}
\usepackage{nicefrac}
\usepackage{microtype}
\usepackage{xcolor}

\usepackage{graphicx}
\usepackage{subcaption}
\usepackage{amsmath}
\usepackage{amssymb}
\usepackage{mathtools}
\usepackage{amsthm}
\usepackage{multirow}
\usepackage[capitalize,noabbrev]{cleveref}

\theoremstyle{plain}

\theoremstyle{definition}

\theoremstyle{remark}

\title{MHWA: Multi-timescale\\
Hierarchical World-Action Model}

\author{%
  Pengcheng Pan$^{1,2,4}$\thanks{These authors contributed equally.}
  \quad Guoqing Ma$^{1,3,4}$\footnotemark[1]
  \quad Yuhan Zhang$^{1,2,4}$
  \quad Yang Chen$^{1,4}$ \\
  \bfseries Yichen Liu$^{1,3,4}$
  \quad Ziheng Li$^{1,2,4}$
  \quad Shan Yu$^{1,4}$\thanks{Corresponding author: \texttt{shan.yu@nlpr.ia.ac.cn}} \\[4pt]
  {\small $^{1}$Laboratory of Brain Atlas and Brain-inspired Intelligence,} \\
  {\small Institute of Automation, Chinese Academy of Sciences, Beijing, China} \\
  {\small $^{2}$School of Artificial Intelligence, University of Chinese Academy of Sciences, Beijing, China} \\
  {\small $^{3}$School of Future Technology, University of Chinese Academy of Sciences, Beijing, China} \\
  {\small $^{4}$State Key Laboratory of Brain Cognition and Brain-inspired Intelligence Technology,} \\
  {\small Institute of Automation, Chinese Academy of Sciences, Beijing, China} \\[3pt]
  {\footnotesize \texttt{\{panpengcheng19, maguoqing22\}@mails.ucas.ac.cn}} \\
  {\footnotesize \texttt{\{zhangyuhan2022, yang.chen, liuyichen2024, liziheng2024\}@ia.ac.cn}}
}

\begin{document}

\maketitle

\begin{abstract}
  Autoregressive world models have emerged as a powerful paradigm for sample efficient decision making, yet their open loop rollouts often exhibit increasing drift that leads to error accumulation even within a relatively small planning horizon.
  Hierarchical temporal abstraction is a natural candidate to mitigate such drift by reducing the effective prediction horizon; however, existing hierarchical designs can introduce added structural and conceptual complexity, sometimes making them difficult to control and limiting gains over flat baselines.
  To this end, we introduce MHWA (Multi-timescale Hierarchical World-Action model), which employs a strided high level context channel to reduce global context update frequency, effectively curbing the propagation of compounding drift. 
  Crucially, via context aware routing, the gating network dynamically reconfigures a low level mixture of experts by using both global context and current observations, thereby enabling flexible high--low coordination, helping alleviate interference commonly observed in static hierarchies.
  Empirically, using RLiable's seed-by-game aggregation, MHWA (77M parameters) achieves an IQM human normalized score (HNS) of 0.884 across three training seeds on a 14-game Atari subset using 10\% subsampled offline data, outperforming a matched JOWA-70M baseline (0.529) and achieving numerically comparable performance to JOWA-150M while using nearly half the parameters and 70.88\% fewer training FLOPs per sequence.
  
\end{abstract}

\section{Introduction}
\label{sec:intro}

Model-Based Reinforcement Learning (MBRL) offers a promising avenue for sample efficient learning by enabling agents to plan within a learned World Model (WM)~\cite{ha2018world, hafner2023dreamerv3}. 
However, maintaining predictive consistency under multi-step open loop rollouts remains a fundamental challenge.
State-of-the-art approaches predominantly employ autoregressive Transformer architectures~\cite{robine2023iris, micheli2022iris, cheng2024scaling}.
While effective at short range prediction, autoregressive rollouts can exhibit increasing drift with rollout depth: once the model conditions on its own imperfect predictions, small local errors may compound under distribution shift, causing imagined trajectories to deviate from the real dynamics~\cite{talvitie2017self, asadi2019compounding}.

A prevalent response to this challenge is to scale model capacity, since larger models can better absorb heterogeneity and noise. 
A prominent example is the Jointly-Optimized World-Action (JOWA) model, which scales up to 150 million parameters to achieve strong performance in multi-game offline reinforcement learning~\cite{cheng2024scaling}. 
While effective, such brute force scaling can be computationally costly, requiring substantial GPU resources and long training cycles, which raises the barrier to iteration.

Hierarchical Reinforcement Learning (HRL) provides an alternative design axis by introducing temporal abstraction~\cite{sutton1999between, hafner2022deep}. 
This motivates the emerging line of Hierarchical Model-Based RL (HMBRL), or Hierarchical World Models (HWM)~\cite{schiewer2024exploring}, which aim to separate slow varying context from fast local dynamics and thereby reduce the effective horizon over which global consistency must be maintained. 
In principle, such multi-timescale structure could improve longer range credit assignment and planning while retaining the sample efficiency benefits of world models.

However, the HWM landscape remains relatively sparse compared to model free approaches; as noted by \citet{schiewer2024exploring}, combining hierarchy and world modeling often yields systems that are ``highly complex and difficult to control,'' and the added hierarchy does not reliably outperform flat baselines.
A key reason is that existing designs face two coupled forms of interference.
\\

\textbf{Conceptual (semantic) interference.}
In many goal conditioned hierarchical formulations, higher-level decisions (e.g., subgoals) are evaluated through the evolving competence of the lower-level goal-reaching policy. As multiple levels learn simultaneously, the semantic meaning and value of a high-level decision can drift over training, inducing a \emph{moving target} learning signal and exacerbating non-stationarity \citep{pateria2021hierarchical, schiewer2024exploring}. This instability is especially pronounced when the abstract interface is high-dimensional or when the hierarchy relies on objectives whose targets implicitly change as the lower level improves.

\textbf{Structural interference.}
A second issue is architectural: many hierarchical agents rely on a single universal low-level network conditioned on goals, forcing one set of parameters to fit heterogeneous behaviors and dynamics regimes. Such heavy parameter sharing can entangle incompatible modes and lead to gradient competition across subtasks \citep{pateria2021hierarchical}. This motivates modular parameterizations that can specialize while remaining scalable.

Together, these observations suggest that without explicit decoupling and flexible coordination, simply adding hierarchy can introduce complexity that becomes a burden rather than a benefit.

To address these challenges, we propose the \textbf{M}ulti-timescale \textbf{H}ierarchical \textbf{W}orld \textbf{A}ction model (\textbf{MHWA}), a problem structured hierarchical world action model for reducing compounding errors. MHWA treats hierarchy not as a standalone remedy, but as a design that must explicitly control the interference it introduces: slow context inference is decoupled from native resolution local dynamics, and low-level computation is dynamically routed through context aware experts.

\textbf{Scope.}
We treat multi-timescale abstraction and context driven routing as an engineering hypothesis for reducing open loop drift under heterogeneous multi-game dynamics, and provide a supporting error bound intuition in Appendix~\ref{app:error_bound}. 
Our claims are scoped to multi-game offline Atari, which we use as a controlled benchmark for testing whether structural inductive bias can improve rollout stability without relying solely on brute force scaling.
\\

Our contributions are summarized as follows:

\textbf{An interference aware hierarchy for reducing compounding errors.}
We propose MHWA, a problem structured hierarchical world action model that treats hierarchy not as a standalone remedy, but as a mechanism that must be carefully decoupled. 
Our central hypothesis is that temporal abstraction can mitigate compounding errors in autoregressive world models when conceptual interference between hierarchy levels and structural interference across dynamics modes are explicitly addressed. 
MHWA implements this idea by coupling slow context inference with native resolution local dynamics through a strided High-Level Manager and a context routed Low-Level Worker.

\textbf{Isolated slow context learning for conceptual interference.}
The High-Level Manager learns a slow varying latent context using a skip step predictive objective ($\mathcal{L}_{high}$), while gradients from low-level losses are blocked before reaching the manager. This prevents the high-level context from being driven by low-level noise or moving target optimization signals.

\textbf{Context aware expert routing for structural interference.}
The Low-Level Worker replaces a monolithic feed forward parameterization with a sparsely activated MoE. Its gate conditions jointly on the broadcast high-level context and the instantaneous token state, enabling expert selection to be both mode aware and locally reactive under heterogeneous multi-game dynamics.

\textbf{Structure versus scale under a controlled benchmark.}
On a 14-game Atari subset with 10\% subsampled offline data, MHWA (77M) achieves an IQM HNS of 0.884 across three training seeds, outperforming JOWA-70M (0.529) trained on the same data with the same batch setting. Its performance is numerically comparable to the published JOWA-150M result (0.833)~\citep{cheng2024scaling}, with nearly half the parameters and 70.88\% fewer forward+backward FLOPs per training sequence.

\section{Related Work}
\label{sec:related_work}

\textbf{Offline RL, world models, and compounding errors.}
Offline RL learns policies from fixed datasets, where distribution shift between the behavior data and the learned policy is a central challenge~\citep{levine2020offline_survey, fujimoto2019bcq}. Model-free methods often mitigate this issue through policy constraints or conservative value estimation~\citep{fujimoto2021td3bc, kumar2020conservative, kostrikov2021iql}, while model-based methods learn dynamics models for planning or policy improvement~\citep{kidambi2020mopo, yu2020moorel}. World models provide a natural interface for model-based decision making~\citep{ha2018worldmodels, hafner2019dream}. Recent sequence modeling approaches, including Decision Transformer, IRIS, and JOWA, cast RL as autoregressive prediction over observation-action histories~\citep{chen2021dt, micheli2022iris, cheng2024scaling}. However, when such models roll out open loop and condition on their own predictions, local errors can compound over time. MHWA targets this failure mode by introducing temporal abstraction into tokenized world action modeling.

\textbf{Temporal abstraction and conceptual interference.}
Hierarchical RL decomposes behavior across temporal scales through options, managers, or subgoals~\citep{sutton1999between, kulkarni2016hdqn, bacon2017option, vezhnevets2017feudal, nachum2018data, zhang2026diffusion}. Hierarchical world models similarly aim to separate slow varying context from fast local dynamics~\citep{schiewer2024exploring, shaj2023multi, saxena2021clockwork}. This separation is appealing for reducing effective rollout horizon, but it can also introduce conceptual interference: high-level decisions may be evaluated through a lower-level policy whose competence and semantics evolve during training~\citep{pateria2021hierarchical, schiewer2024exploring}. THICK~\citep{gumbsch2023learning} learns temporal abstraction from sparse latent context changes and predicts future latent scenarios associated with those changes. In contrast, MHWA uses a fixed strided manager with skip step observation token supervision and stop-gradient-isolated context broadcast, avoiding direct dependence of the high-level context on low-level optimization signals.
Beyond temporal abstraction, perturbation-based local learning has also been explored to improve training scalability and convergence efficiency in spiking neural networks~\citep{MA2026133139}.

\textbf{Modularity and structural interference.}
A second challenge is structural interference, where heterogeneous behaviors or dynamics regimes are forced through a heavily shared low-level parameterization~\citep{pateria2021hierarchical}. Sparse Mixture-of-Experts models address a related problem by routing inputs to a subset of specialized parameters~\citep{shazeer2017outrageously, fedus2022switch, riquelme2021vmoe}. MoE world models support modular dynamics modeling for multi-task and multi-agent learning and planning~\citep{tang2025moe,zhao2025learning}. MHWA differs by coupling sparse routing with an explicit high-level context channel: the worker gate conditions jointly on the broadcast slow context and the instantaneous token state, enabling expert selection to be both mode aware and locally reactive. Thus, we do not claim skip step prediction, hierarchy, or MoE routing as standalone novelties; our contribution is an interference aware integration of these mechanisms for reducing drift in tokenized multi-game offline world-action modeling.

\begin{figure*}[t]
    \centering
    \includegraphics[width=0.80\textwidth]{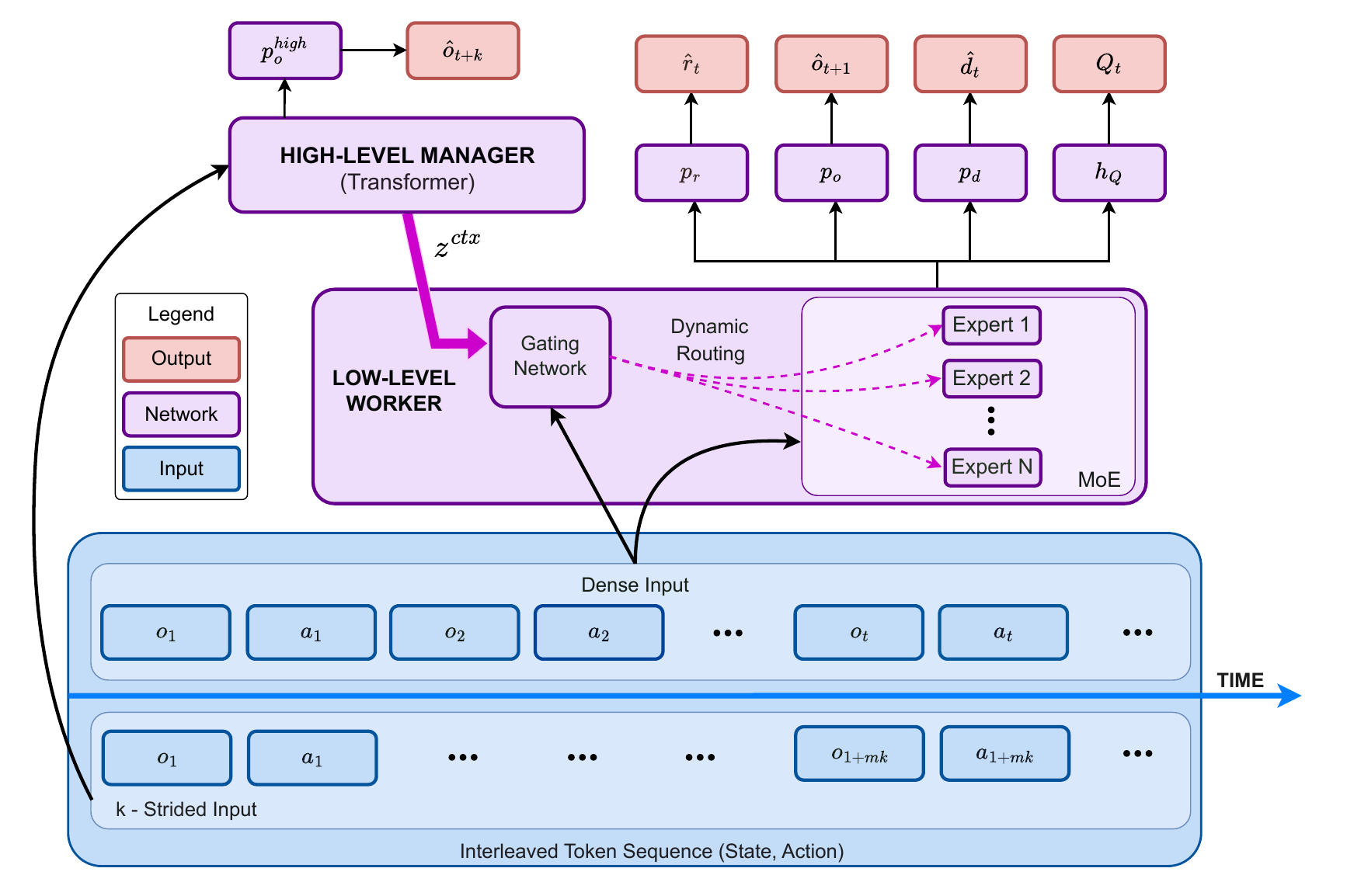}
    \caption{\textbf{The architecture of MHWA.} It is illustrated at a context-update timestep $t = 1 + m \cdot k$. 
    The model processes an interleaved sequence of observation ($o_t$) and action ($a_t$) tokens.\\
    \textbf{(Top)} At this specific strided step, the High-Level Manager consumes the downsampled history to infer a slowly varying global context $z^{ctx}$, supervised by a skip-step prediction head $p^{high}_o$ that forecasts the observation at the next coarse step $\hat{o}_{t+k}$. We refer to the original, complete sequence as the dense trajectory, in contrast to the downsampled input used by the manager.\\
    \textbf{(Bottom)} The Low-Level Worker operates at the atomic resolution. It employs a Context-Aware Dynamic Routing mechanism where a Gating Network utilizes both the current dense input and the refreshed high-level context $z^{ctx}$ to dynamically route computation to a sparse subset of experts (MoE).
    The worker's output drives the world model heads ($p_o, p_r, p_d$) and the critic head ($h_Q$).}
    \label{fig:architecture}
\end{figure*}

\section{The Multi-timescale Hierarchical World-Action Model (MHWA)}
\label{sec:method}

We propose the \textbf{M}ulti-timescale \textbf{H}ierarchical \textbf{W}orld \textbf{A}ction model (\textbf{MHWA}), an asymmetric dual-layer architecture for reducing compounding errors in tokenized autoregressive world models. 
MHWA targets two interference modes: conceptual interference between hierarchy levels, and structural interference across heterogeneous dynamics modes. 
It addresses the former by isolating a slow high-level context through gradient detachment, and the latter by using context-aware sparse routing in the low-level dynamics model.

\subsection{Problem Formulation}

We consider offline reinforcement learning as a Markov Decision Process (MDP) $\mathcal{M} = \langle \mathcal{S}, \mathcal{A}, \mathcal{P}, \mathcal{R}, \gamma \rangle$. 
In visual RL, the agent observes high-dimensional observations $o_t \in \mathcal{O}$ rather than ground-truth states, and we treat the stacked observation history as the state representation. 
Given a static dataset $\mathcal{D} = \{ \tau_i \}_{i=1}^N$, where $\tau = (o_1, a_1, r_1, \dots, o_T, a_T, r_T)$, the goal is to learn predictive transition and outcome models that support long-term return maximization.

Following tokenized world-action models, each observation is encoded into $K$ discrete visual tokens using a VQ-VAE tokenizer, reusing the pretrained tokenizer from JOWA. 
These visual tokens are interleaved with executed action tokens to form a single autoregressive sequence. 
Token, positional, and task embeddings are combined before both Transformers, retaining JOWA's explicit task conditioning.
An autoregressive predictor processes this sequence and produces embeddings consumed by four heads: 
(i) a next-observation token head $p_o$, 
(ii) a reward head $p_r$, 
(iii) a termination head $p_d$, and 
(iv) a Q-value head $h_Q$ trained with offline TD learning. 
In MHWA, this predictor is implemented by the Low-Level Worker modulated by the High-Level Manager, as described in Sec.~\ref{subsec:architecture}. 
We refer to the components producing $(p_o,p_r,p_d)$ as the \emph{world-part} and those producing $h_Q$ as the \emph{action-part}; both world-modeling and TD/CQL critic losses back-propagate through the autoregressive predictor.

Standard autoregressive world models treat trajectories as uniform token sequences, where prediction errors may accumulate with rollout depth. 
MHWA instead introduces temporal abstraction to infer global context less frequently, while preserving local dynamics prediction at the native token resolution.

\subsection{Asymmetric Dual-Layer Architecture}
\label{subsec:architecture}

MHWA is organized around the two interference modes above. 
To mitigate \emph{conceptual interference}, the high-level representation should capture slow-varying context without being dominated by low-level reconstruction noise. 
To mitigate \emph{structural interference}, the low-level model should avoid fitting heterogeneous dynamics with a single shared feed-forward parameterization. 
Accordingly, MHWA retains the tokenized world-action interface and standard prediction heads, while replacing the monolithic autoregressive predictor with an asymmetric two-level hierarchy (Figure~\ref{fig:architecture}): 
(i) a High-Level Manager that infers slowly varying context at a reduced update frequency, and 
(ii) a Low-Level Worker implemented as an autoregressive Transformer whose feed-forward sublayers are replaced by sparse Mixture-of-Experts (MoE) layers. 
The worker output drives the world-model heads $(p_o,p_r,p_d)$ and the critic head $(h_Q)$. 
The manager provides an isolated slow context channel, while the worker combines this context with the instantaneous state for dynamic expert routing. 
This couples temporal abstraction with modular local prediction, directly targeting conceptual and structural interference in heterogeneous multi-game dynamics.

\subsection{High-Level Manager: Context Inference}

The High-Level Manager targets conceptual interference by learning a slow context representation on a strided temporal scale. 
Let $x_{t} \in \mathbb{R}^{d}$ denote the input embedding at atomic step $t$, where $x_t$ represents either an observation ($o_t$) or an action ($a_t$) embedding because the input sequence interleaves observations and actions. 
We introduce a high-level index $m$, where each step corresponds to atomic timestamp $t = 1 + m \cdot k$ with stride factor $k$; we set $k=4$.
At coarse step $m$, the manager consumes the strided token sequence $\mathbf{X}^{(m)}_{slow} = (x_{1}, x_{1+k}, \dots, x_{1+m \cdot k})$, a $k$-fold temporal downsampling of the atomic sequence. 
A causal Transformer encoder~\cite{vaswani2017attention} computes:
\begin{equation}
    h^{high}_{m} = \text{Transformer}_{High}(\mathbf{X}^{(m)}_{slow})
\end{equation}
From $h^{high}_{m}$, a trainable Context Projector produces a continuous context vector $z^{ctx}_{m}=\mathrm{Projector}(\mathrm{stopgrad}(h^{high}_{m}))$, which parametrizes the slow-varying context of the current dynamics regime rather than acting as a discrete subgoal.

\textbf{Context Broadcasting and Gradient Isolation.}
We broadcast $z^{ctx}_{m}$ to the Low-Level Worker and hold it constant for the subsequent window of $k$ atomic steps. 
Stop-gradient is placed between the manager and the Context Projector, leaving the projector, router, and worker trainable under the low-level objectives.
Thus, the manager is trained through its own auxiliary objective instead of low-level gradients, preventing short-range prediction noise from overwriting the high-level context. 
This is the main mechanism for controlling conceptual interference between hierarchy levels.

\textbf{Implementation of Stride.}
Temporal abstraction is implemented as a fixed token-level stride on the interleaved sequence. 
This matches the tokenized world-action interface, where visual and action tokens are flattened into one stream. 
The stride shortens the manager sequence for efficient slow-context inference, while the Low-Level Worker still handles fine-grained native-resolution prediction. 
Although striding discards some tokens, VQ-VAE visual codes contain sufficient redundancy for coarse context inference.

\textbf{Distinction from Static Execution.}
The stride $k$ sets the manager's temporal sampling and prediction scale; the context-refresh interval $r$ is set to $k$ by default.
Unlike static temporal abstractions~\cite{schiewer2024exploring} that lock low-level policies or dynamics for fixed intervals, MHWA keeps the low-level routing mechanism operating at the atomic timescale $t$. 
The worker therefore remains locally reactive by conditioning on both the broadcast context and the instantaneous state.

\subsection{Low-Level Worker: Context-Aware Dynamic Routing}
\label{subsec:routing}

The Low-Level Worker targets structural interference. 
In heterogeneous multi-game dynamics, a monolithic low-level FFN must model diverse visual patterns, action effects, and transition modes with shared parameters, which can entangle incompatible behaviors. 
MHWA replaces these FFN sublayers with sparse MoE layers~\cite{shazeer2017outrageously, fedus2022switch}. 
The routing mechanism conditions on both the broadcast high-level context and the instantaneous token state, allowing expert selection to reflect slow dynamics modes while remaining locally reactive.

Let $x_t$ denote the intermediate token representation in the residual stream at token position $t$, and let $z^{ctx}_m$ be the broadcast high-level context for $m=\lfloor (t-1)/k \rfloor$. 
A gating projection $W_g\in\mathbb{R}^{2d\times N}$ maps the concatenated input to routing logits:
\begin{equation}
l_t=(x_t\oplus z^{ctx}_m)W_g,
\end{equation}
where $\oplus$ denotes concatenation and $N$ is the number of experts $\{E_j\}_{j=1}^{N}$. 
We apply Top-$K$ sparse routing by selecting $\mathcal{T}_t=\mathrm{TopK}(l_t,K_{\mathrm{route}})$ and computing masked Softmax weights. 
We set  $N=4$; $K_{\mathrm{route}}=2$:
\begin{equation}
w_{t,j}=\frac{\exp(l_{t,j})}{\sum_{r\in\mathcal{T}_t}\exp(l_{t,r})},\; j\in\mathcal{T}_t;\qquad w_{t,j}=0,\; j\notin\mathcal{T}_t.
\end{equation}
The MoE output is:
\begin{equation}
y_t=x_t+\sum_{j\in\mathcal{T}_t} w_{t,j}\,E_j(x_t).
\end{equation}
Conditioning the gate on both $z^{ctx}_m$ and $x_t$ makes routing both mode-aware and locally reactive: the context term encourages specialization across slow-varying dynamics regimes, while the state term preserves sensitivity to instantaneous visual and action-dependent details. 
This addresses structural interference by replacing a single shared low-level transformation with sparse, context-conditioned expert computation. 
We use no capacity factor, token dropping, or truncation; all selected tokens are computed, and expert collapse is mitigated with an auxiliary load-balancing regularizer. We provide complete architectural and routing implementation details in Appendix~\ref{app:architecture_details}.

\subsection{Learning Objectives}
\label{subsec:objectives}

We train the MHWA architecture jointly but with cross-level stop-gradient.  
Accordingly, the total objective function comprises \textbf{four distinct components}. 
The first two ($\mathcal{L}_{base}, \mathcal{L}_{critic}$) are adopted from baseline JOWA~\cite{cheng2024scaling}.
The latter two ($\mathcal{L}_{high}, \mathcal{L}_{aux}$) are novel structural constraints introduced by MHWA, specifically designed to enforce hierarchical representation learning and dynamic expert specialization.

\textbf{1. Base World Model Objectives ($\mathcal{L}_{base}$):} 
Following standard autoregressive approaches, the Low-Level Worker transforms the input sequence into a final hidden representation $h_t$. This representation is then used to predict the next discrete observation token $o_{t+1}$, reward $r_t$, and termination signal $d_t$ via standard Cross-Entropy loss. This ensures the model accurately simulates the environment's local dynamics:
\begin{equation}
\begin{aligned}
    \mathcal{L}_{base} = -\sum_{t} \Big[ & \underbrace{\log p_o(o_{t+1} | h_t)}_{\text{Observation}}  + \underbrace{\log p_r(r_t | h_t)}_{\text{Reward}} + \underbrace{\log p_d(d_t | h_t)}_{\text{Termination}} \Big]
\end{aligned}
\end{equation}

\textbf{2. Offline Critic Objectives ($\mathcal{L}_{critic}$):} 
For policy evaluation, we employ a distributional Q-function learned via Temporal Difference (TD) learning, augmented with a Conservative Q-Learning (CQL) penalty to mitigate value overestimation in offline regimes~\cite{kumar2020conservative}. The critic loss is defined as:
\begin{equation}
    \mathcal{L}_{critic} = \mathcal{L}_{TD}(\theta_Q) + \alpha_{cql} \mathcal{L}_{CQL}(\theta_Q)
\end{equation}
where $\mathcal{L}_{TD}$ minimizes the distributional Bellman error (using C51) and $\mathcal{L}_{CQL}$ penalizes Q-values on out-of-distribution actions.

\textbf{3. High-Level Representation Learning ($\mathcal{L}_{high}$):} 
A core challenge in hierarchical modeling is preventing the high-level representation from collapsing into a trivial identity mapping of low-level features or overfitting to transient, hard-to-predict details. 
To encourage the High-Level Manager to distill slow-varying and semantically meaningful context, we introduce a skip-step predictive objective. 
Concretely, an MLP projection head $p_{high}$ predicts the token at the next sampled position using only the corresponding high-level hidden state $h^{high}_{m}$. 
This skip-step gap acts as an information bottleneck, discouraging reliance on ephemeral noise and promoting representations that remain predictive across multiple token steps:
\begin{equation}
    \mathcal{L}_{high} = -\sum_{m} \log p_o^{high}(o_{(m+1)k} \mid h^{high}_{m})
\end{equation}

\textbf{4. Routing Regularization ($\mathcal{L}_{aux}$):} 
To prevent expert collapse and ensure balanced resource utilization, we minimize the Kullback-Leibler (KL) divergence between the batch-wise expert selection frequency $\hat{p}$ and the uniform distribution $\mathcal{U}$:
\begin{equation}
    \mathcal{L}_{aux} = D_{KL}(\hat{p} \parallel \mathcal{U}) = \sum_{i=1}^{N_{exp}} \hat{p}_i \log \hat{p}_i + \text{const}
\end{equation}
where $\hat{p}_i = \frac{1}{B \cdot T \cdot K_{\mathrm{route}}} \sum_{b,t} \mathbb{I}(i \in \mathcal{T}_{b,t})$ represents the normalized selection ratio of expert $i$ within a batch, here $B$ denotes the batch size, $T$ denotes the number of routed token positions per sequence, $K_{\mathrm{route}}$ is the number of active experts per token. 
Minimizing this objective maximizes the entropy of the routing distribution, enforcing a balanced load across the expert bank.

\textbf{Total Objective:} The final training objective aggregates the main generative tasks with these structural constraints:
\begin{equation}
    \mathcal{L}_{total} = \underbrace{\mathcal{L}_{base} + \mathcal{L}_{critic}}_{\text{Main Task}} + \underbrace{\lambda_{h} \mathcal{L}_{high}}_{\text{Representation}} + \underbrace{\lambda_{a} \mathcal{L}_{aux}}_{\text{Regularization}}
\end{equation}
where $\lambda_{h}$ and $\lambda_{a}$ are hyperparameters governing the strength of the hierarchical and sparsity constraints, respectively. We set $\lambda_{h}=1$, $\lambda_{a}=0.1$.

\begingroup
\setlength{\parskip}{4pt}
\setlength{\textfloatsep}{12pt plus 2pt minus 2pt}
\setlength{\floatsep}{8pt plus 2pt minus 2pt}
\setlength{\intextsep}{8pt plus 2pt minus 2pt}
\section{Experiment}

\subsection{Experimental Setup}

\textbf{Dataset.} 
While previous multi-game offline RL works such as JOWA utilized the DQN Replay Dataset \cite{agarwal2020optimistic}, the original source is no longer publicly available. Following the recommendation of the original authors, we utilize the \texttt{rlu-atari} version of the dataset released as part of the RL Unplugged benchmark \cite{gulcehre2020rl}, which maintains the identical data distribution (50M transitions per game per run). However, since the \texttt{rlu-atari} benchmark excludes the game \textit{Berzerk}, we select a subset of 14 games for training, maintaining the difficulty distribution (easy, medium, hard) \cite{gulcehre2020rl}, see Appendix~\ref{app:games}. 
Consistent with \cite{lee2022multi,kumar2022offline,cheng2024scaling}, we use data from 2 out of 5 training runs. Following the JOWA baseline \cite{cheng2024scaling}, we uniformly draw 10\% of transitions at random, resulting in approximately 10M transitions (including frame stacking) per game.

\textbf{Training.} 
MHWA has 77.27M total parameters, with 60.49M parameters activated per forward pass. 
It uses $K=36$ visual tokens per frame and adopts the same $L=8$-frame context window as JOWA~\cite{cheng2024scaling}.
We train on 2 NVIDIA A40 GPUs with AdamW~\cite{loshchilov2017decoupled} and a learning rate of $1\times10^{-4}$. 
To match the baseline token exposure under constrained compute, we use batch size 1024 and 0.875M gradient steps, compared with the published baseline batch size 512 and 1.75M steps. 
Thus, the total number of processed interaction tokens remains matched. 
MHWA requires 21.6 GPU-days on 2 A40 GPUs.
We compare training computation using forward+backward FLOPs with identical inputs: 8 frames, 296 tokens per sequence, and batch size 256 (Table~\ref{tab:matched_comparison}). MHWA uses 42.78\% fewer FLOPs per sequence than JOWA-70M and 70.88\% fewer than JOWA-150M; the counting details are given in Appendix~\ref{app:training_compute}.

\textbf{Evaluation and Metrics.} 
We follow the Atari evaluation procedure of JOWA~\cite{cheng2024scaling}, reporting the average return over 16 episodes per game.
Performance is measured by Human Normalized Score (HNS), $\frac{\text{score}-\text{score}_{\text{random}}}{\text{score}_{\text{human}}-\text{score}_{\text{random}}}$. 
For aggregate comparison, we report IQM and Median HNS following RLiable~\cite{agarwal2021deep}.

\begin{table*}[t]
\centering
\caption{\textbf{Performance comparison on 14 Atari games.} Baseline scores are from~\citet{cheng2024scaling}. We report raw scores for each game and aggregate metrics (Median and IQM Human-Normalized Scores) at the bottom. The best raw score for each game, excluding reference scores, is highlighted in \textbf{bold}. ``Ours'' denotes our 77M-parameter MHWA model. For Ours, per-game raw scores are averaged over three independently trained seeds; bottom aggregate rows report mean$\pm$sample standard deviation over seed-level aggregate metrics. Full per-seed results are provided in Appendix~\ref{app:three_seed_results}.}
\label{tab:main_results}
\resizebox{\textwidth}{!}{%
\begin{tabular}{l|rr|rrrr|rrr|r}
\toprule
\textbf{Game} & \multicolumn{1}{c}{\textbf{Random}} & \multicolumn{1}{c|}{\textbf{Human}} & \multicolumn{1}{c}{\textbf{MTBC}} & \multicolumn{1}{c}{\textbf{MGDT}} & \multicolumn{1}{c}{\textbf{EDT}} & \multicolumn{1}{c|}{\textbf{Scaled-QL}} & \multicolumn{3}{c|}{\textbf{JOWA}} & \multicolumn{1}{c}{\textbf{Ours}} \\
 & & & \multicolumn{1}{c}{120M} & \multicolumn{1}{c}{200M} & \multicolumn{1}{c}{200M} & \multicolumn{1}{c|}{80M} & \multicolumn{1}{c}{40M} & \multicolumn{1}{c}{70M} & \multicolumn{1}{c|}{150M} & \multicolumn{1}{c}{77M} \\
\midrule
Assault & 222.4 & 742.0 & 1203.5 & 1741.5 & 1915.5 & 2292.0 & 1578.3 & 1402.8 
& \textbf{2302.0} & 1068.8 \\
Atlantis & 12850.0 & 29028.1 & 37812.5 & 2565750.0 & 2108166.7 & 49100.0 & 41662.5 & 228825.0 
& \textbf{2690387.5} & 1690733.3 \\
BeamRider & 363.9 & 16926.5 & 786.0 & 6011.3 & 4149.8 & \textbf{7023.5} & 880.3 & 854.4 
& 3498.0 & 4202.1 \\
Carnival & 0.0 & 3800.0 & \textbf{5560.0} & 2610.0 & 3678.8 & 1940.0 & 4685.0 & 3812.1 
& 5316.0 & 881.7 \\
Centipede & 2090.9 & 12017.0 & 4046.6 & 4604.0 & 3389.9 & 3650.3 & \textbf{5669.3} & 4485.0 
& 4677.0 & 4970.9 \\
ChopperCommand & 811.0 & 7387.8 & 256.3 & 3300.8 & 3443.8 & 853.8 & 3118.8 & 2568.8 
& 3812.5 & \textbf{4547.9} \\
DemonAttack & 152.1 & 1971.0 & 2611.3 & 6549.4 & 3455.7 & \textbf{7936.5} & 1233.8 & 1409.4 
& 3547.8 & 6593.5 \\
NameThisGame & 2292.3 & 8049.0 & 6045.0 & 6610.5 & 7060.0 & 6461.7 & 7335.6 & 7507.1 
& 11421.0 & \textbf{13280.0} \\
Phoenix & 761.4 & 7242.6 & 1446.9 & 5120.5 & 5320.0 & 4961.7 & 1608.8 & 5070.0 
& \textbf{5348.0} & 4476.1 \\
Seaquest & 68.4 & 42054.7 & 120.0 & 2720.0 & \textbf{3160.4} & 650.0 & 1033.1 & 1040.0 
& 2725.0 & 1215.0 \\
SpaceInvaders & 148.0 & 1668.7 & 605.0 & 742.5 & 513.8 & 714.4 & 694.7 & 805.3 
& 744.7 & \textbf{2948.3} \\
StarGunner & 664.0 & 10250.0 & 1493.8 & 8625.5 & 9550.0 & 4728.6 & 5737.5 & \textbf{21700.0} 
& 18150.0 & 19100.0 \\
TimePilot & 3568.0 & 5229.2 & 762.5 & 3866.7 & 2812.5 & 4072.7 & 3268.8 & 1100.3 
& 3669.0 & \textbf{4700.0} \\
Zaxxon & 32.5 & 9173.3 & 0.0 & 462.5 & 325.5 & 0.0 & 0.0 & 200.0 
& \textbf{2163.0} & 1166.7 \\
\midrule
\#Superhuman & 0 & - & 4 & 3 & 3 & 3 & 3 & 4 & \textbf{6} & 5.7$\pm$0.6 \\
Median HNS & 0 & 1.000 & 0.151 & 0.532 & 0.552 & 0.413 & 0.360 & 0.549 & 0.582 & \textbf{0.676$\pm$0.049} \\
IQM HNS & 0 & 1.000 & 0.292 & 0.537 & 0.554 & 0.443 & 0.397 & 0.531 & 0.833 & \textbf{0.867$\pm$0.047} \\
\bottomrule
\end{tabular}%
}
\end{table*}

\subsection{Overall Performance Comparison}

We compare our method against strong multi-task offline RL baselines, including MTBC, MGDT \cite{lee2022multi}, EDT \cite{wu2023elastic}, Scaled-QL \cite{kumar2022offline},  and the SOTA model-based approach, JOWA \cite{cheng2024scaling}. 
The baseline results in Table~\ref{tab:main_results} are sourced from the original JOWA paper \cite{cheng2024scaling} under the same 10\% low-data protocol on the 14 shared games.
For MHWA, Table~\ref{tab:main_results} reports the mean and sample standard deviation of seed-level IQMs, while Table~\ref{tab:matched_comparison} reports IQM aggregated across the seed-by-game score matrix.
We additionally train JOWA-70M with three independent seeds on the same 14-game \texttt{rlu-atari} dataset and with the same batch setting as MHWA.
Using RLiable's seed-by-game aggregation~\cite{agarwal2021deep}, MHWA-77M achieves IQM HNS \textbf{0.884} with a 95\% stratified-bootstrap confidence interval of [0.736, 1.031], compared with 0.529 [0.459, 0.602] for the matched JOWA-70M baseline (Table~\ref{tab:matched_comparison}).
This comparison demonstrates the benefit of the hierarchical routing architecture at comparable model capacity. MHWA is also numerically comparable to the published JOWA-150M result (0.833).

Under the JOWA held-out fine-tuning protocol, MHWA also obtains 0.643$\pm$0.007 IQM DNS on five unseen Atari games, close to JOWA-150M (0.647) and above JOWA-70M (0.603); see Appendix~\ref{app:ood_finetuning}.


\begin{table}[t]
\centering
\caption{\textbf{Matched performance and training computation.} JOWA-70M and MHWA use three training seeds on the same 14-game dataset and batch setting. We report seed-by-game IQM HNS and 95\% stratified-bootstrap CIs. JOWA-150M performance is the published reference. Forward+backward FLOPs use identical input shapes.}
\label{tab:matched_comparison}
\small
\begin{tabular}{lccc}
\toprule
Model & IQM HNS & 95\% CI & GFLOPs/sequence \\
\midrule
JOWA-70M & 0.529 & [0.459, 0.602] & 84.238 \\
MHWA-77M & \textbf{0.884} & [0.736, 1.031] & \textbf{48.199} \\
\midrule
JOWA-150M~\citep{cheng2024scaling} & 0.833 & -- & 165.524 \\
\bottomrule
\end{tabular}
\end{table}

\subsection{Open-loop Dynamics Modeling}

To evaluate error accumulation under autoregressive rollout, we measure open-loop reconstruction quality over steps $t=5\ldots 8$ within the context window. 
For each game, we sample held-out trajectory segments disjoint from the training data, condition each model on the first four ground-truth frames and corresponding actions, and then roll out from $t=5$ without injecting ground-truth observations. 
Actions are kept fixed to the dataset actions, so differences in MSE/PSNR/SSIM against ground-truth frames reflect dynamics-modeling errors rather than policy differences.

As shown in Figure~\ref{fig:reconstruction_trend}, the monolithic JOWA-70M baseline shows a marked deterioration as rollout depth increases, indicating that reducing capacity without architectural changes amplifies error accumulation even within the context window. 
In contrast, MHWA-77M curbs error propagation and achieves predictive stability comparable to JOWA-150M. 
A longer-horizon diagnostic from $t=5$ to $t=54$ further shows that MHWA outperforms the smaller monolithic baseline and remains comparable to the larger baseline; see Appendix~\ref{app:long_open_loop}.

Overall, these results suggest that MHWA improves open-loop stability through structural inductive bias rather than relying solely on brute-force scaling.

\begin{figure*}[t]
    \centering
    \includegraphics[width=0.95\textwidth]{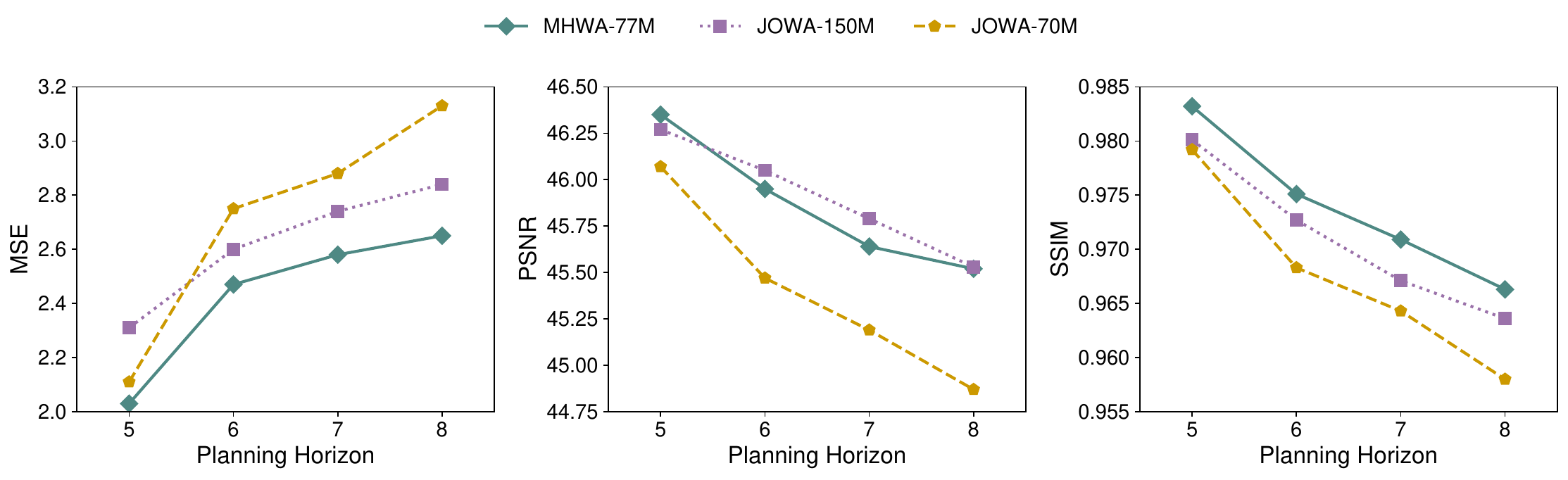}
    \caption{\textbf{Quantitative comparison of open-loop prediction within the context window ($T=5$ to $8$).}
    We analyze reconstruction quality (MSE, PSNR, SSIM) of MHWA-77M against JOWA-150M and a parameter-matched JOWA-70M.
    While the reduced-capacity JOWA-70M degrades rapidly as rollout depth increases, MHWA remains stable and, in terms of MSE and SSIM, matches or exceeds the larger JOWA-150M.
    The gap between MHWA and JOWA-70M is consistent with the benefit of structural inductive bias under constrained parameters.}
    \label{fig:reconstruction_trend}
\end{figure*}

\subsection{Ablation Studies}
\label{subsec:ablation}

We evaluate design choices on a fixed 6-game subset using a 44M MHWA variant (Appendix~\ref{app:games}). 
This controlled ablation setting follows the subset-ablation protocol used by JOWA~\cite{cheng2024scaling} and keeps the study computationally feasible under our available budget. 
For each variant, we first average the 16 evaluation episodes within each game to obtain one HNS per game, and then compute IQM and Median HNS over the six per-game scores. 
Table~\ref{tab:ablation_horizontal} reports the resulting aggregate scores, together with the relative IQM drop compared to the full model. 
We provide detailed per-game raw scores in Appendix~\ref{app:ablation_per_game}. 
To isolate architectural effects from capacity, \textit{w/o MoE} widens the dense FFN and \textit{MoE-only (w/o Manager)} widens the experts to keep parameters at $\approx$44M.

\textbf{Temporal abstraction.} 
The stride ablations show that the high-level context update frequency is a key factor. 
Collapsing the hierarchy to $k=1$ causes the largest degradation among all variants, with a 62.1\% relative IQM drop, indicating that temporal abstraction is central to MHWA's benefit. 
The default $k=4$ performs best in this 44M/6-game setting, while $k=2$ and $k=8$ remain lower but competitive, with relative IQM drops of 20.9\% and 8.3\%, respectively. 
This suggests that insufficient abstraction is especially harmful, whereas overly coarse abstraction has a milder effect in this setting.

\textbf{Routing mechanism.} 
The \textit{MoE-only (w/o Manager)} variant removes the manager, Context Projector, skip-step head, and $\mathcal{L}_{high}$, retaining the task embedding and low-level MoE; its IQM is 0.445. Removing the state input to the router yields IQM 0.359.
The larger drop from removing state routing (36.3\%) suggests that local state information is especially important for reactive fine-grained prediction. 
The 21.1\% drop from removing the manager pathway further indicates that trajectory-dependent context helps expert selection across slow-varying dynamics modes.

\textbf{MoE modularity.} 
Replacing the sparse MoE with a dense FFN at matched parameter count (\textit{w/o MoE}) reduces IQM to 0.450, corresponding to a relative IQM drop of 20.2\%. 
This is consistent with modular specialization being beneficial under heterogeneous multi-game dynamics.

\textbf{High-level supervision and routing regularization.} 
Removing the skip-step objective $\mathcal{L}_{high}$ yields a moderate drop (IQM: $0.564 \rightarrow 0.503$, 10.8\%), suggesting that explicit high-level supervision helps the manager learn stride-level predictive context rather than short-range noisy features. 
Removing the load-balancing regularizer $\mathcal{L}_{aux}$ causes a larger drop (IQM: $0.564 \rightarrow 0.444$, 21.3\%), supporting its role in preventing expert collapse and maintaining stable specialization.

\textbf{Gradient isolation.}
The completed factorial ablation (Table~\ref{tab:gradient_factorial}) shows that stop-gradient improves IQM both with $\mathcal{L}_{high}$ ($0.533\rightarrow0.564$) and without it ($0.387\rightarrow0.503$).
Without $\mathcal{L}_{high}$, the manager has fixed weights but supplies trajectory-dependent hidden states to the trainable Context Projector. Replacing those states with zeros or Gaussian noise reduces IQM from 0.503 to 0.406 or 0.381 (Appendix~\ref{app:context_controls}).

\textbf{Expert configuration and context refresh.}
The default four-expert, Top-2 configuration performs best among the tested alternatives; four-expert Top-1 obtains 0.542 and eight-expert Top-2 obtains 0.511 (Appendix~\ref{app:expert_configuration}).
Separating manager stride from context refresh, retraining with $k=4,r=2$ gives IQM 0.558 versus 0.564 for the default $r=4$, with Median HNS 0.484 versus 0.463 (Appendix~\ref{app:context_refresh}).

\begin{table}[t]
\centering
\caption{\textbf{High-level supervision and gradient isolation.} IQM HNS under the 44M/6-game ablation protocol. The Context Projector and worker remain trainable in all four settings.}
\label{tab:gradient_factorial}
\small
\begin{tabular}{lcc}
\toprule
 & Stop-gradient on & Stop-gradient off \\
\midrule
$\mathcal{L}_{high}$ on & \textbf{0.564} & 0.533 \\
$\mathcal{L}_{high}$ off & 0.503 & 0.387 \\
\bottomrule
\end{tabular}
\end{table}

\begin{table*}[t]
    \centering
    \caption{\textbf{Ablation analysis on the 6-game subset (44M).} 
    For each variant, we average 16 evaluation episodes within each game and report Median and IQM human-normalized score (HNS) over the six resulting per-game scores. 
    $\Delta_{\mathrm{IQM}}(\%)$ denotes the relative IQM drop compared to the full model, i.e., $(\mathrm{IQM}_{\mathrm{Full}}-\mathrm{IQM}_{\mathrm{variant}})/\mathrm{IQM}_{\mathrm{Full}}\times 100$.
    Bold and underline mark the largest and second-largest $\Delta_{\mathrm{IQM}}(\%)$, respectively.
    All variants are controlled to have $\approx$44M parameters for fair comparison. 
    Detailed per-game raw scores are provided in Appendix~\ref{app:ablation_per_game}.}
    \label{tab:ablation_horizontal}
    \resizebox{\textwidth}{!}{%
    \begin{tabular}{lccccccccc}
        \toprule
        \textbf{Metric} 
        & \textbf{Full} 
        & \textbf{w/o $\mathcal{L}_{high}$} 
        & \textbf{w/o $\mathcal{L}_{aux}$} 
        & \textbf{w/o MoE} 
        & \textbf{MoE-only} 
        & \textbf{w/o State} 
        & \textbf{$k=1$} 
        & \textbf{$k=2$} 
        & \textbf{$k=8$} \\
        \midrule
        \textbf{Median HNS} 
        & 0.463
        & 0.425
        & 0.439
        & 0.412
        & 0.386
        & 0.364
        & 0.183
        & 0.446
        & 0.469 \\
        \textbf{IQM HNS} 
        & 0.564
        & 0.503
        & 0.444
        & 0.450
        & 0.445
        & 0.359
        & 0.214
        & 0.446
        & 0.517 \\
        \textbf{$\Delta_{\mathrm{IQM}}(\%)$} 
        & --
        & 10.8
        & 21.3
        & 20.2
        & 21.1
        & \underline{36.3}
        & \textbf{62.1}
        & 20.9
        & 8.3 \\
        \bottomrule
    \end{tabular}%
    }
\end{table*}

\begin{figure*}[!htbp]
    \centering
    \begin{subfigure}{0.32\textwidth}
        \includegraphics[width=\linewidth]{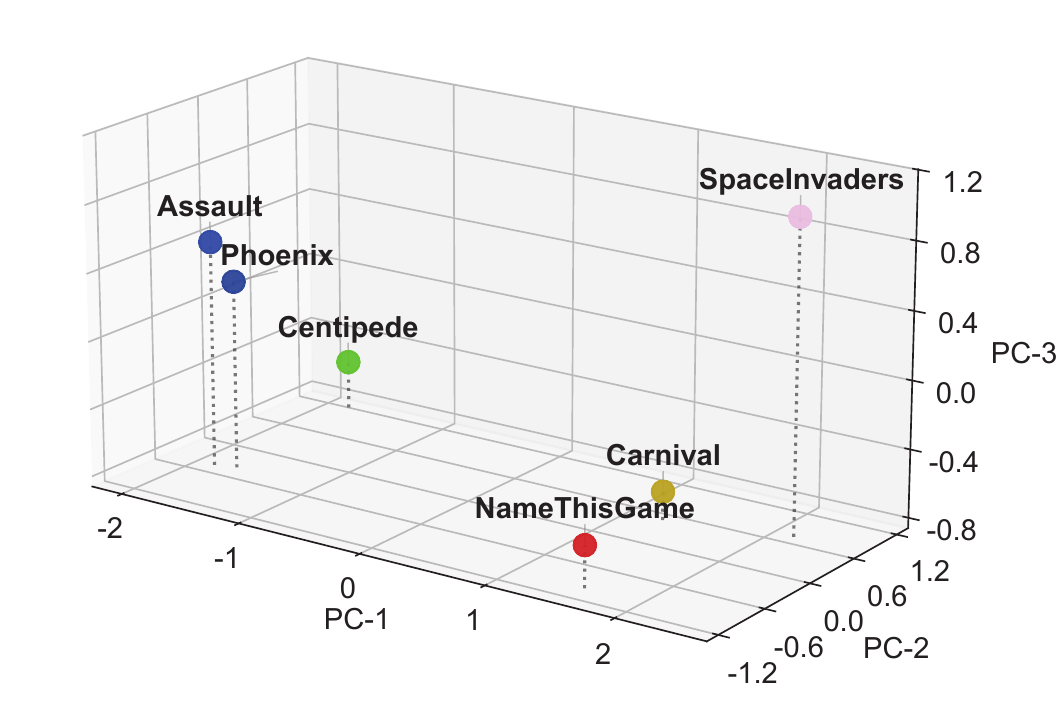}
        \caption{Full MHWA}
    \end{subfigure}
    \hfill
    \begin{subfigure}{0.32\textwidth}
        \includegraphics[width=\linewidth]{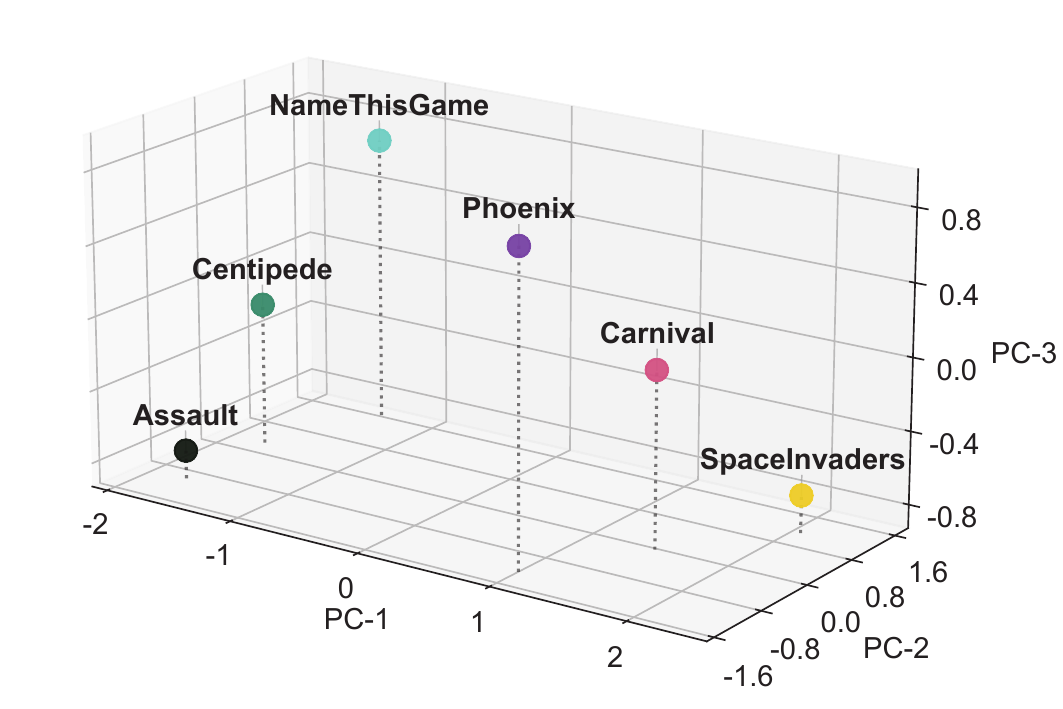}
        \caption{w/o High-Level Loss}
    \end{subfigure}
    \hfill
    \begin{subfigure}{0.32\textwidth}
        \includegraphics[width=\linewidth]{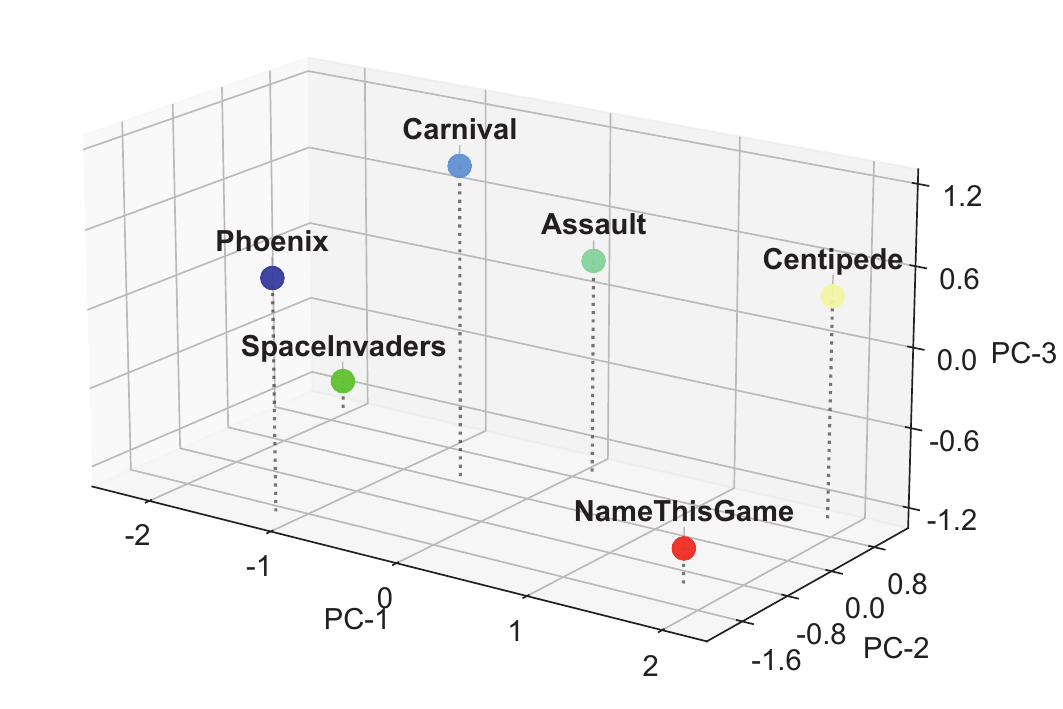}
        \caption{w/o Stride ($k=1$)}
    \end{subfigure}
    \caption{\textbf{PCA visualization of expert activations in the last MoE layer.} 
    (a) Full MHWA maps \textit{Assault} and \textit{Phoenix} closely together, indicating they utilize similar expert combinations, which aligns with their shared vertical shooting mechanics. 
    (b, c) This structural similarity vanishes without high-level supervision or temporal stride, indicating a failure to capture shared cross-game dynamics.}
    \label{fig:pca_visualization}
\end{figure*}

\subsection{Interpretability of Representation Learning}
\label{subsec:interpretability}

To probe whether MHWA captures semantically meaningful dynamics, we visualize the PCA-projected expert activation frequencies of the last MoE layer. \textit{Assault} and \textit{Phoenix} are both vertical space shooters sharing identical mechanics, and the projection shows that this similarity is reflected in the full MHWA model (Fig.~\ref{fig:pca_visualization}a) via highly similar expert-usage patterns. In contrast, this correspondence is not preserved when removing high-level supervision (Fig.~\ref{fig:pca_visualization}b) or the temporal stride (Fig.~\ref{fig:pca_visualization}c), suggesting that the $\mathcal{L}_{high}$ and time abstraction design is necessary for the model to recover such task-level commonalities from expert activations.

\section{Discussion and Conclusion}
\label{sec:discussion_conclusion}

We introduced MHWA, a multi-timescale hierarchical world-action model that combines a strided, gradient-isolated high-level manager with a context-aware MoE low-level worker. The manager reduces how often global context must be inferred, while the worker remains reactive at the native token resolution and routes computation through specialized experts. On the 10\% offline Atari benchmark, MHWA outperforms a matched three-seed JOWA-70M baseline, achieves numerically comparable performance to JOWA-150M with 77M parameters and 70.88\% fewer training FLOPs per sequence, and shows competitive adaptation to held-out Atari games.

The current evidence is intentionally scoped. We evaluate primarily in discrete-action offline Atari with a pretrained tokenizer and fixed stride, and extending to adaptive temporal abstraction~\cite{neitz2018adaptive}, continuous control, and broader domains remains future work.


\clearpage
\endgroup
\bibliographystyle{unsrtnat}
\bibliography{ref}

\newpage
\appendix
\section{Atari Game Subset}
\label{app:games}

We follow the game difficulty taxonomy introduced in the RL Unplugged Atari benchmark~\cite{gulcehre2020rl} and report the properties of the games used in our multi-game training set.
Compared to the 20-game suite reported in prior work, we do \emph{not} include games used for fine-tuning/transfer evaluation, and we additionally exclude \textit{Berzerk} because it is not available in the \texttt{rlu-atari} release.
The resulting training set contains 14 Atari games, listed in Table~\ref{tab:games} along with their difficulty labels and action-space sizes.

\begin{table}[htbp]
    \centering
    \caption{Atari games used in our multi-game training set, with difficulty labels from~\cite{gulcehre2020rl} and discrete action-space sizes.}
    \label{tab:games}
    \begin{tabular}{lccc}
        \toprule
        Game & Difficulty & $|\mathcal{A}|$ & Split \\
        \midrule
        Assault & Medium & 7 & Train \\
        Atlantis & Hard & 18 & Train \\
        BeamRider & Medium & 9 & Train \\
        Carnival & Medium & 6 & Train \\
        Centipede & Medium & 18 & Train \\
        ChopperCommand & Easy & 18 & Train \\
        DemonAttack & Easy & 6 & Train \\
        NameThisGame & Easy & 6 & Train \\
        Phoenix & Easy & 8 & Train \\
        Seaquest & Easy & 18 & Train \\
        SpaceInvaders & Easy & 6 & Train \\
        StarGunner & Medium & 18 & Train \\
        TimePilot & Easy & 10 & Train \\
        Zaxxon & Easy & 18 & Train \\
        \bottomrule
    \end{tabular}
\end{table}

\paragraph{6-game subset for ablations.}
Following the subset protocol used in JOWA, we conduct ablations on a fixed 6-game subset:
\textit{Assault}, \textit{Carnival}, \textit{Centipede}, \textit{NameThisGame}, \textit{Phoenix}, and \textit{SpaceInvaders}.
This subset is chosen to keep a reasonably balanced mix of difficulty levels and discrete action-space sizes.

\section{Architecture and Parameterization Details}
\label{app:architecture_details}

The High-Level Manager is also a 4-layer, 8-head, 512-dimensional Transformer, but uses a dense MLP with expansion ratio 4 instead of MoE. Its output is detached before a trainable $512 \rightarrow 512$ Context Projector and broadcast across the following stride window. Low-level losses train the Context Projector, router, and worker, while $\mathcal{L}_{high}$ trains the manager. Both Transformers receive the shared token, positional, and task embeddings. The tokenized interface uses 37 tokens per block, consisting of 36 visual tokens and one action token, with 8 blocks per sequence. The critic head uses one Q-head, MLP dimension 768, and dropout 0.01.

The Low-Level Worker is a 4-layer causal Transformer with 8 attention heads and embedding dimension 512. Each feed-forward sublayer is replaced by an MoE layer with $N=4$ experts and Top-$K=2$ routing. Each expert is a $512 \rightarrow 2048 \rightarrow 512$ GELU MLP. The routing gate is a bias-free linear projection over the concatenation of the current token state and the broadcast high-level context. We do not use token dropping or capacity truncation; all selected tokens are computed, and exactly two experts are active per token.

The full MHWA model contains 77.27M parameters, with approximately 60.49M parameters activated per forward pass including the tokenizer.

\subsection{Training Computation}
\label{app:training_compute}

We count forward+backward FLOPs for the main experimental configurations using 8 frames and 296 tokens per sequence, with batch size 256. The count includes the complete forward graph, world-model, critic, and auxiliary losses, and backward computation. For MHWA, it includes both hierarchy levels, all four MoE layers, four stored experts with Top-2 routing, load balancing, and high-level observation prediction. MoE computation is counted for the activated experts.

\begin{table}[htbp]
\centering
\caption{\textbf{Forward+backward training computation at matched input shape.}}
\label{tab:training_compute}
\begin{tabular}{lrr}
\toprule
Model & GFLOPs/sequence & TFLOPs/training step \\
\midrule
MHWA-77M & 48.199 & 12.339 \\
JOWA-70M & 84.238 & 21.565 \\
JOWA-150M & 165.524 & 42.374 \\
\bottomrule
\end{tabular}
\end{table}

\section{A Simple Error-Propagation View}
\label{app:error_bound}

We provide a simple formal view of when a strided context model can have a smaller one-step open-loop rollout-error bound than a flat model. This analysis is intended to isolate the core mechanism of MHWA, rather than to provide a general performance guarantee. We focus on open-loop state prediction under a fixed action sequence.

Let the latent state be $x_t=(s_t,c_t)$, where $s_t\in\mathcal S$ denotes the fast state and $c_t\in\mathcal C$ denotes the slow context. Let $\mathcal A$ be any metric action space, either discrete or continuous. The true dynamics are
\begin{equation}
    s_{t+1}=f(s_t,c_t,a_t).
\end{equation}
In MHWA, at macro-step $m$, the manager predicts a context $\hat c_m$, which is held fixed for the next $k$ local steps, while the worker rolls out
\begin{equation}
    \hat s_{t+1}=\hat f(\hat s_t,\hat c_m,\tilde a_t),
\end{equation}
where $\tilde a_t$ denotes the action representation used by the model.

\paragraph{Assumptions.}
We use the following standard assumptions for open-loop dynamics prediction.

\textbf{Lipschitz continuity.}
There exist constants $L_s,L_c,L_a\ge 0$ such that
\begin{equation}
\label{eq:lipschitz_dynamics}
d_{\mathcal S}\!\left(f(s,c,a),f(s',c',a')\right)
\le
L_s d_{\mathcal S}(s,s')
+
L_c d_{\mathcal C}(c,c')
+
L_a d_{\mathcal A}(a,a'),
\end{equation}
where $d_{\mathcal S}$, $d_{\mathcal C}$, and $d_{\mathcal A}$ are metrics on the fast-state, context, and action spaces.

\textbf{Slow context drift.}
There exists $\rho\ge 0$ such that
\begin{equation}
\label{eq:slow_context_drift}
d_{\mathcal C}(c_t,c_{t+1})\le \rho.
\end{equation}
Therefore, for a local step $t=mk+j$ with $j\in\{0,\ldots,k-1\}$,
\begin{equation}
\label{eq:context_drift_within_window}
d_{\mathcal C}(c_{mk},c_t)\le j\rho.
\end{equation}

\textbf{Manager context error.}
At each macro-step $m$, the manager's context prediction satisfies
\begin{equation}
\label{eq:manager_context_error}
d_{\mathcal C}(c_{mk},\hat c_m)\le \beta_H.
\end{equation}

\textbf{Local one-step modeling and interface error.}
Assume the local dynamics model has one-step approximation error
\begin{equation}
\label{eq:local_model_error}
d_{\mathcal S}\!\left(f(s,c,a),\hat f(s,c,a)\right)\le \varepsilon_f.
\end{equation}
We further collect action-representation and latent/interface approximation errors into a single quantity $\eta$:
\begin{equation}
\label{eq:eta_definition}
d_{\mathcal S}\!\left(f(s,c,a),\hat f(s,c,\tilde a)\right)\le \eta,
\qquad
\eta := \varepsilon_f + L_a\delta_a + \delta_x,
\end{equation}
where $L_a\delta_a$ bounds the error induced by action approximation and $\delta_x$ collects additional latent or interface approximation error. In continuous-action or continuous-latent settings, these terms need not arise from discretization and may be smaller when no discretization interface is used.

\paragraph{Derivation.}
Define the open-loop rollout error
\begin{equation}
    e_t := d_{\mathcal S}(s_t,\hat s_t).
\end{equation}
For a local step $t=mk+j$, we have
\begin{align}
e_{t+1}
&=
d_{\mathcal S}(s_{t+1},\hat s_{t+1}) \nonumber \\
&=
d_{\mathcal S}\!\left(
f(s_t,c_t,a_t),
\hat f(\hat s_t,\hat c_m,\tilde a_t)
\right).
\end{align}
By the triangle inequality,
\begin{align}
e_{t+1}
&\le
d_{\mathcal S}\!\left(
f(s_t,c_t,a_t),
f(\hat s_t,\hat c_m,a_t)
\right)
+
d_{\mathcal S}\!\left(
f(\hat s_t,\hat c_m,a_t),
\hat f(\hat s_t,\hat c_m,\tilde a_t)
\right).
\label{eq:triangle_error}
\end{align}
Applying the Lipschitz condition in Eq.~\eqref{eq:lipschitz_dynamics} to the first term gives
\begin{align}
d_{\mathcal S}\!\left(
f(s_t,c_t,a_t),
f(\hat s_t,\hat c_m,a_t)
\right)
&\le
L_s d_{\mathcal S}(s_t,\hat s_t)
+
L_c d_{\mathcal C}(c_t,\hat c_m).
\end{align}
Using the local error definition in Eq.~\eqref{eq:eta_definition}, Eq.~\eqref{eq:triangle_error} yields
\begin{equation}
\label{eq:error_before_context_bound}
e_{t+1}
\le
L_s e_t
+
L_c d_{\mathcal C}(c_t,\hat c_m)
+
\eta.
\end{equation}
It remains to bound the context mismatch. By the triangle inequality, slow context drift, and manager context error,
\begin{align}
d_{\mathcal C}(c_t,\hat c_m)
&\le
d_{\mathcal C}(c_t,c_{mk})
+
d_{\mathcal C}(c_{mk},\hat c_m) \nonumber \\
&\le
j\rho+\beta_H.
\label{eq:context_mismatch_bound}
\end{align}
Substituting Eq.~\eqref{eq:context_mismatch_bound} into Eq.~\eqref{eq:error_before_context_bound}, we obtain the MHWA one-step open-loop error bound
\begin{equation}
\label{eq:mhwa_error_bound}
e_{t+1}
=
e_{mk+j+1}
\le
L_s e_t+\eta+L_c(\beta_H+j\rho).
\end{equation}

\paragraph{Comparison to a flat model.}
For a controlled comparison, we match the local one-step error $\eta$ and isolate the effect of context inference. Let a flat model re-infer context at every step, with per-step context error bounded by $\beta_F$. Its analogous error recursion is
\begin{equation}
\label{eq:flat_error_bound}
e^F_{t+1}
\le
L_s e^F_t+\eta+L_c\beta_F.
\end{equation}
For matched current rollout error $e_t=e_t^F$, MHWA has a one-step upper bound no larger than the flat model's whenever
\begin{equation}
\label{eq:local_condition}
\beta_H+j\rho \le \beta_F.
\end{equation}
Since the worst case within a stride window is $j=k-1$, a sufficient condition over the whole window is
\begin{equation}
\label{eq:window_condition}
\beta_H+(k-1)\rho \le \beta_F.
\end{equation}
This condition identifies when the strided context yields a one-step error bound no larger than the flat model's.

\paragraph{Implications for transfer across settings.}
The comparison above does not rely on discrete actions or tokenized latent interfaces: discrete-action Atari is one special case, while continuous actions and continuous latent interfaces are covered by the same metric-space formulation, with interface effects absorbed into $\eta$. Likewise, in this narrow architectural sense, the comparison does not depend on whether the model is trained offline or online; it depends on the achieved error terms $(\beta_H,\beta_F,\eta,\rho)$ rather than on how they are learned.

\newpage
\section{Per-seed Results on the 14-game Benchmark}
\label{app:three_seed_results}

For each seed and game, we average 16 evaluation episodes before computing HNS. Tables~\ref{tab:three_seed_results} and~\ref{tab:jowa_seed_variation} report per-seed raw returns and aggregate metrics. Mean and Sample SD summarize the three seed-level values in each row. The three MHWA seed-level IQMs are 0.837, 0.922, and 0.844. RLiable aggregation of the $3\times14$ seed-by-game score matrix yields IQM 0.884 with a 95\% stratified-bootstrap CI of [0.736, 1.031].

\begin{table}[htbp]
\centering
\caption{\textbf{Per-seed results of MHWA-77M on the 14-game Atari benchmark.}}
\label{tab:three_seed_results}
\renewcommand{\arraystretch}{0.95}
\resizebox{\textwidth}{!}{%
\begin{tabular}{lrrrrr}
\toprule
\textbf{Game} & \textbf{Seed 0} & \textbf{Seed 1} & \textbf{Seed 2} & \textbf{Mean} & \textbf{Sample SD} \\
\midrule
Assault & 1,209.3 & 1,109.1 & 888.1 & 1,068.8 & 164.3 \\
Atlantis & 3,320,456.3 & 1,078,587.5 & 673,156.2 & 1,690,733.3 & 1,425,865.1 \\
BeamRider & 4,077.6 & 4,920.0 & 3,608.8 & 4,202.1 & 664.4 \\
Carnival & 1,300.0 & 305.0 & 1,040.0 & 881.7 & 516.1 \\
Centipede & 5,689.3 & 3,705.4 & 5,518.1 & 4,970.9 & 1,099.3 \\
ChopperCommand & 3,300.0 & 5,793.8 & 4,550.0 & 4,547.9 & 1,246.9 \\
DemonAttack & 8,036.3 & 6,766.6 & 4,977.5 & 6,593.5 & 1,536.7 \\
NameThisGame & 15,750.0 & 9,140.0 & 14,950.0 & 13,280.0 & 3,607.6 \\
Phoenix & 5,218.1 & 4,503.8 & 3,706.3 & 4,476.1 & 756.3 \\
Seaquest & 1,683.8 & 926.3 & 1,035.0 & 1,215.0 & 409.6 \\
SpaceInvaders & 3,350.0 & 3,495.0 & 2,000.0 & 2,948.3 & 824.5 \\
StarGunner & 7,800.0 & 24,000.0 & 25,500.0 & 19,100.0 & 9,814.8 \\
TimePilot & 4,500.0 & 4,700.0 & 4,900.0 & 4,700.0 & 200.0 \\
Zaxxon & 700.0 & 2,800.0 & 0.0 & 1,166.7 & 1,457.2 \\
\midrule
\#Superhuman & 5 & 6 & 6 & 5.7 & 0.6 \\
Median HNS & 0.624 & 0.720 & 0.685 & 0.676 & 0.049 \\
IQM HNS & 0.837 & 0.922 & 0.844 & 0.867 & 0.047 \\
\bottomrule
\end{tabular}
}
\end{table}

\subsection{Matched JOWA-70M Results}

JOWA-70M is trained with three independent seeds on the same 14-game \texttt{rlu-atari} dataset and with the same batch setting as MHWA. Its seed-by-game IQM is 0.529 with a 95\% stratified-bootstrap CI of [0.459, 0.602]. Table~\ref{tab:jowa_seed_variation} reports the complete per-seed results and game-level cross-seed variation, complementing the aggregate comparison in Table~\ref{tab:matched_comparison}.

\begin{table}[htbp]
\centering
\caption{\textbf{Per-seed results of matched JOWA-70M on the 14-game Atari benchmark.}}
\label{tab:jowa_seed_variation}
\renewcommand{\arraystretch}{0.95}
\resizebox{\textwidth}{!}{%
\begin{tabular}{lrrrrr}
\toprule
\textbf{Game} & \textbf{Seed 0} & \textbf{Seed 1} & \textbf{Seed 2} & \textbf{Mean} & \textbf{Sample SD} \\
\midrule
Assault & 1,408.3 & 858.1 & 1,208.2 & 1,158.2 & 278.4 \\
Atlantis & 3,167,275.0 & 125,787.5 & 2,885,637.5 & 2,059,566.7 & 1,680,611.9 \\
BeamRider & 853.5 & 608.5 & 915.0 & 792.3 & 162.1 \\
Carnival & 3,243.8 & 1,293.8 & 690.0 & 1,742.5 & 1,334.7 \\
Centipede & 8,937.4 & 4,619.2 & 4,102.1 & 5,886.2 & 2,655.0 \\
ChopperCommand & 675.0 & 681.3 & 737.5 & 697.9 & 34.4 \\
DemonAttack & 8,020.3 & 7,030.4 & 6,190.9 & 7,080.6 & 915.7 \\
NameThisGame & 8,199.4 & 6,298.8 & 7,550.6 & 7,349.6 & 966.1 \\
Phoenix & 3,906.3 & 6,133.8 & 4,850.0 & 4,963.3 & 1,118.1 \\
Seaquest & 237.5 & 235.0 & 433.8 & 302.1 & 114.0 \\
SpaceInvaders & 993.1 & 1,495.0 & 1,640.0 & 1,376.0 & 339.4 \\
StarGunner & 8,468.8 & 10,887.5 & 10,593.8 & 9,983.3 & 1,319.9 \\
TimePilot & 825.0 & 1,993.8 & 887.5 & 1,235.4 & 657.5 \\
Zaxxon & 318.8 & 1,368.8 & 0.0 & 562.5 & 716.2 \\
\midrule
\#Superhuman & 4 & 4 & 4 & 4.0 & 0.0 \\
Median HNS & 0.623 & 0.518 & 0.417 & 0.519 & 0.103 \\
IQM HNS & 0.565 & 0.528 & 0.495 & 0.529 & 0.035 \\
\bottomrule
\end{tabular}
}
\end{table}

\section{Held-out Game Fine-tuning}
\label{app:ood_finetuning}

We evaluate transfer to unseen games using the held-out Atari fine-tuning protocol used by JOWA. The held-out games are \textit{Gravitar}, \textit{MsPacman}, \textit{Pong}, \textit{Robotank}, and \textit{YarsRevenge}. Each game is fine-tuned separately using 5k expert-level transitions drawn from the last 20\% of DQN-Replay. To match JOWA's data exposure, which uses batch size 32 for 50k updates, we use batch size 1024 for 1562 updates, yielding 1,599,488 samples versus 1.60M samples for JOWA.

\begin{table}[htbp]
\centering
\caption{\textbf{Held-out Atari fine-tuning results.}
We report per-game raw scores and aggregate DQN-normalized scores (DNS). MHWA is evaluated over three seeds; the MHWA Mean column averages raw scores across the three seeds for per-game rows and reports mean $\pm$ sample standard deviation over seed-level aggregate DNS for the bottom rows.}
\label{tab:ood_finetuning}
\resizebox{\textwidth}{!}{%
\begin{tabular}{lrr|rr|rrrr}
\toprule
\textbf{Game} 
& \textbf{Random} 
& \textbf{DQN} 
& \textbf{JOWA-70M} 
& \textbf{JOWA-150M} 
& \textbf{MHWA Seed 0} 
& \textbf{MHWA Seed 1} 
& \textbf{MHWA Seed 2} 
& \textbf{MHWA Mean} \\
\midrule
Gravitar     & 173.0  & 473.0   & 387.4   & 273.3   & 375.2   & 271.8   & 284.4   & 310.5 \\
MsPacman     & 307.3  & 3085.6  & 908.0   & 2016.7  & 1437.5  & 2006.2  & 1971.0  & 1804.9 \\
Pong         & -20.7  & 19.5    & 14.6    & 17.7    & 17.0    & 18.1    & 18.1    & 17.7 \\
Robotank     & 2.2    & 63.9    & 13.8    & 25.0    & 12.4    & 11.3    & 9.3     & 11.0 \\
YarsRevenge  & 3092.9 & 18089.9 & 16339.2 & 17506.2 & 16056.0 & 17760.4 & 19280.6 & 17699.0 \\
\midrule
Mean DNS     & -- & -- & 0.576 & 0.647 & 0.610 & 0.606 & 0.626 & 0.614$\pm$0.010 \\
Median DNS   & -- & -- & 0.715 & 0.615 & 0.674 & 0.611 & 0.599 & 0.628$\pm$0.040 \\
IQM DNS      & -- & -- & 0.603 & 0.647 & 0.648 & 0.635 & 0.645 & 0.643$\pm$0.007 \\
\bottomrule
\end{tabular}%
}
\end{table}

Across the three MHWA seeds, the IQM DNS values are 0.648, 0.635, and 0.645, yielding 0.643$\pm$0.007. MHWA is competitive on \textit{Pong} and \textit{YarsRevenge}, remains reasonably close on \textit{MsPacman} and \textit{Gravitar}, and is weaker on \textit{Robotank}, suggesting that the transfer benefit is meaningful but not uniform across all held-out games. As a sanity check, we also ran one more expensive fine-tuning run with batch size 32, which reached IQM DNS 0.668; since this result is available for only one seed, we do not use it as the main comparison.

\section{Per-game Ablation Results}
\label{app:ablation_per_game}

\begin{table}[htbp]
\centering
\caption{\textbf{Per-game raw scores for ablation variants on the 6-game subset.}
All variants use the 44M controlled setting. For each game and variant, we report the raw score averaged over 16 evaluation episodes. The bottom rows report Median and IQM HNS over the six per-game scores, matching Table~\ref{tab:ablation_horizontal}.}
\label{tab:ablation_per_game}
\resizebox{\textwidth}{!}{%
\begin{tabular}{l|rr|rrrrrrrrr}
\toprule
\textbf{Game}
& \textbf{Random}
& \textbf{Human}
& \textbf{Full}
& \textbf{w/o $\mathcal{L}_{high}$}
& \textbf{w/o $\mathcal{L}_{aux}$}
& \textbf{w/o MoE}
& \textbf{MoE-only}
& \textbf{w/o State}
& \textbf{$k=1$}
& \textbf{$k=2$}
& \textbf{$k=8$} \\
\midrule
Assault 
& 222.4 & 742.0 
& 1086.0 & 989.9 & 930.1 & 917.3 & 902.7 & 745.8 & 1003.6 & 843.2 & 1043.6 \\
Carnival 
& 0.0 & 3800.0 
& 626.7 & 220.0 & 380.0 & 340.0 & 680.0 & 420.0 & 463.8 & 300.0 & 240.0 \\
Centipede 
& 2090.9 & 12017.0 
& 5387.7 & 4527.0 & 4199.0 & 4530.0 & 2982.1 & 2801.3 & 3902.0 & 3228.0 & 3964.0 \\
NameThisGame 
& 2292.3 & 8049.0 
& 9186.5 & 8480.0 & 4160.0 & 6950.0 & 8246.6 & 5690.6 & 3350.0 & 6770.0 & 8270.0 \\
Phoenix 
& 761.4 & 7242.6 
& 3924.0 & 4214.4 & 4350.0 & 2628.1 & 4610.0 & 4040.9 & 1503.1 & 3120.0 & 4300.0 \\
SpaceInvaders 
& 148.0 & 1668.7 
& 813.9 & 630.9 & 1205.0 & 961.9 & 110.0 & 485.0 & 796.9 & 950.0 & 745.0 \\
\midrule
Median HNS
& -- & -- 
& 0.463
& 0.425
& 0.439
& 0.412
& 0.386
& 0.364
& 0.183
& 0.446
& 0.469 \\
IQM HNS
& -- & --
& 0.564
& 0.503
& 0.444
& 0.450
& 0.445
& 0.359
& 0.214
& 0.446
& 0.517 \\
\bottomrule
\end{tabular}%
}
\end{table}

\subsection{Hierarchical Context Controls}
\label{app:context_controls}

The factorial ablation in Table~\ref{tab:gradient_factorial} retains the manager, Context Projector, context-conditioned router, and stride-$k$ broadcasting in all four configurations. With $\mathcal{L}_{high}$ on, the manager is trained by high-level supervision and, when stop-gradient is off, also by low-level objectives. With $\mathcal{L}_{high}$ off, the manager is fixed-weight when stop-gradient is on and trained only by low-level objectives when it is off. The Context Projector, router, and worker remain trainable throughout. MoE-only removes the complete manager pathway and widens the low-level experts to match model size.

We train two additional information controls under the $\mathcal{L}_{high}$-off, stop-gradient-on setting. Both preserve the fixed-weight manager, trainable Context Projector, worker, MoE, parameter count, and $k=4$ broadcasting. They replace the manager hidden state after stop-gradient with zeros or Gaussian noise during both training and evaluation. Zero input produces a learned constant context through the projector bias; Gaussian noise is scaled using batch RMS and varies across context blocks.

\begin{table}[htbp]
\centering
\caption{\textbf{Information controls for the fixed-weight manager pathway.} All variants are separately trained in the 44M/6-game setting with $\mathcal{L}_{high}$ off and stop-gradient on.}
\label{tab:context_controls}
\begin{tabular}{lcc}
\toprule
Hidden state supplied to Context Projector & Median HNS & IQM HNS \\
\midrule
Trajectory-conditioned manager hidden state & \textbf{0.425} & \textbf{0.503} \\
Zero hidden state & 0.424 & 0.406 \\
Gaussian noise & 0.410 & 0.381 \\
\bottomrule
\end{tabular}
\end{table}

The trajectory-conditioned hidden state improves IQM over both controls. Its Median HNS is close to that of the zero-input variant, while IQM captures the larger separation across the six-game score distribution.

\subsection{Expert Number and Routing Sparsity}
\label{app:expert_configuration}

Table~\ref{tab:expert_configuration} varies expert number, Top-$K$, and expert-MLP expansion in the 44M/6-game setting. Let $r_0$ denote the default expansion ratio. The $N=2$, Top-1 configuration doubles the expansion, while $N=8$, Top-4 halves it to keep total expert parameters comparable to the default. The $N=4$, Top-1 variant keeps the default expert bank; the $N=8$, Top-2 variant keeps the default per-expert expansion.

\begin{table}[htbp]
\centering
\caption{\textbf{Expert configuration study.} IQM HNS on the six-game ablation benchmark.}
\label{tab:expert_configuration}
\begin{tabular}{cccc}
\toprule
Experts $N$ & Top-$K$ & MLP expansion & IQM HNS \\
\midrule
2 & 1 & $2r_0$ & 0.429 \\
8 & 4 & $r_0/2$ & 0.488 \\
4 & 1 & $r_0$ & 0.542 \\
8 & 2 & $r_0$ & 0.511 \\
4 & 2 & $r_0$ & \textbf{0.564} \\
\bottomrule
\end{tabular}
\end{table}

Four-expert Top-1 retains 96\% of the default IQM, and the eight-expert alternatives retain 87--91\%. The two-expert configuration shows the largest drop, reflecting the effect of coarser capacity partitioning.

For each configuration, we aggregate final-layer expert activation frequencies for the six games, fit PCA independently, and rank the 15 pairwise distances in the three-dimensional projection. Table~\ref{tab:expert_neighborhoods} reports the rank of the Assault--Phoenix pair. This neighborhood is closest for both Top-2 configurations, while Top-1 reorganizes cross-task sharing. Under four-expert Top-1, Centipede--SpaceInvaders moves from the default's 15th-ranked pair to the 6th-ranked pair; their raw returns are 8950.4 and 917.8, respectively, and the six-game Median HNS is 0.516. For two experts, the normalized usage vector has one effective degree of freedom.

\begin{table}[htbp]
\centering
\caption{\textbf{Routing neighborhoods across expert configurations.} Rank of the Assault--Phoenix distance among 15 game pairs in each configuration's PCA projection; rank 1 is the closest pair.}
\label{tab:expert_neighborhoods}
\begin{tabular}{ccc}
\toprule
Experts $N$ & Top-$K$ & Assault--Phoenix distance rank \\
\midrule
4 & 2 & 1 \\
4 & 1 & 14 \\
8 & 2 & 1 \\
8 & 4 & 4 \\
\bottomrule
\end{tabular}
\end{table}

\subsection{Manager Stride and Context-Refresh Interval}
\label{app:context_refresh}

The manager sampling stride $k$ determines the temporal spacing of its inputs and the high-level prediction scale. The context-refresh interval $r$ determines how often the projected context is refreshed for low-level routing. The default uses $k=r=4$.

We first vary only $r$ at inference using the same checkpoint trained with $k=r=4$. The manager retains its $k=4$ input spacing and maximum context length, with sliding-window refreshes at $r=2$ or $r=1$. We then retrain with $k=4,r=2$ throughout training and evaluation, allowing the router and worker to adapt to the alternative schedule. Table~\ref{tab:context_refresh} separates these results.

\begin{table}[htbp]
\centering
\caption{\textbf{Context-refresh schedules at fixed manager stride $k=4$.} The inference-only variants use the default checkpoint; the retrained variant uses $r=2$ during both training and evaluation.}
\label{tab:context_refresh}
\begin{tabular}{lcccc}
\toprule
Setting & Training $r$ & Evaluation $r$ & Median HNS & IQM HNS \\
\midrule
Default & 4 & 4 & 0.463 & 0.564 \\
Inference-only & 4 & 2 & 0.441 & 0.568 \\
Inference-only & 4 & 1 & 0.441 & 0.533 \\
Retrained & 2 & 2 & 0.484 & 0.558 \\
\bottomrule
\end{tabular}
\end{table}

\begin{table}[htbp]
\centering
\caption{\textbf{Per-game results after retraining with $k=4,r=2$.} Raw returns under the same 44M/6-game protocol.}
\label{tab:context_refresh_per_game}
\begin{tabular}{lrr}
\toprule
Game & Default ($k=4,r=4$) & Retrained ($k=4,r=2$) \\
\midrule
Assault & 1086.0 & 1030.3 \\
Carnival & 626.7 & 758.8 \\
Centipede & 5387.7 & 7129.1 \\
NameThisGame & 9186.5 & 8205.0 \\
Phoenix & 3924.0 & 3751.3 \\
SpaceInvaders & 813.9 & 728.4 \\
\midrule
Median HNS & 0.463 & 0.484 \\
IQM HNS & 0.564 & 0.558 \\
\bottomrule
\end{tabular}
\end{table}

After retraining, $r=2$ improves Carnival and Centipede and lowers returns on the other four games. The resulting aggregate performance is comparable to the default schedule, with a higher Median and a slightly lower IQM. Changing the refresh schedule redistributes game-level performance while preserving the manager's learned temporal scale.

\section{Long-horizon Open-loop Diagnostics}
\label{app:long_open_loop}

\begin{figure}[htbp]
    \centering
    \includegraphics[width=0.90\textwidth]{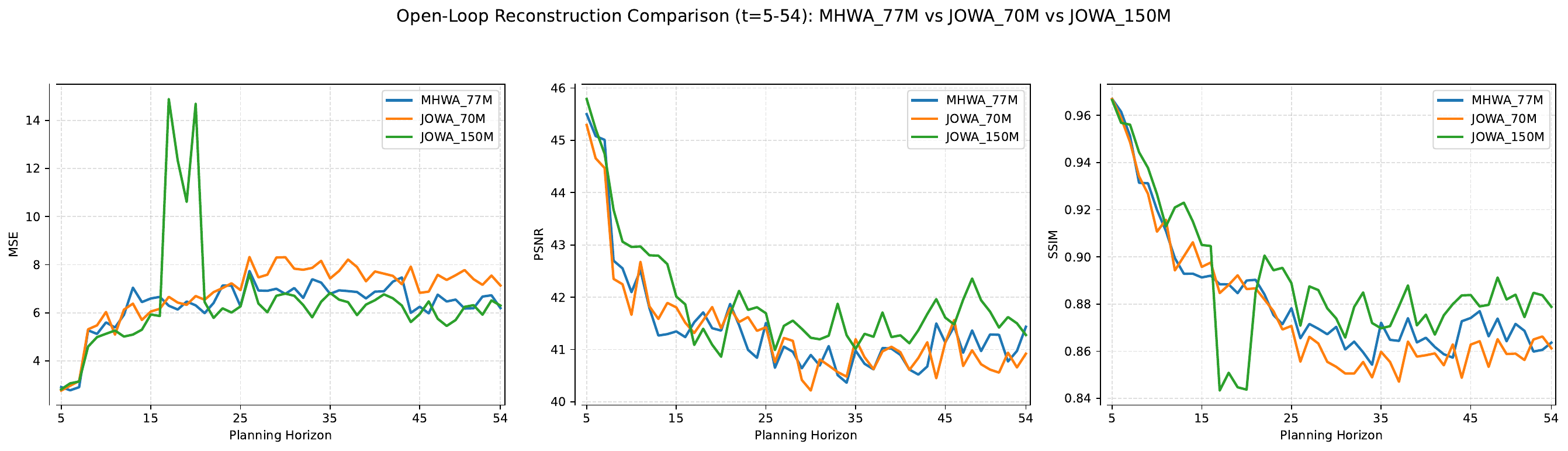}
    \caption{\textbf{Long-horizon open-loop reconstruction diagnostic from $t=5$ to $t=54$.}
    MHWA shows smoother long-horizon degradation than the smaller monolithic baseline and remains broadly comparable to, while being less affected by sharp fluctuations than, the 150M baseline; however, it does not eliminate eventual manifold departure.}
    \label{fig:long_open_loop}
\end{figure}

We further evaluate open-loop rollouts beyond the context window. Starting from four ground-truth context frames, each model autoregressively predicts frames from $t=5$ to $t=54$ while conditioning on the same action sequence. As shown in Figure~\ref{fig:long_open_loop}, the result shows that MHWA exhibits smoother degradation than the smaller monolithic baseline and remains broadly comparable to the 150M baseline. Importantly, the most visible anomaly in JOWA-150M around $t=17$ is not a uniform collapse across games; it is mainly driven by a sharp fluctuation on DemonAttack at $t=17$--$20$, while the same game/timestep does not show a comparable spike for the other models. This diagnostic is intended to test the smoothness of error accumulation rather than to claim indefinitely stable long-horizon generation.

\newpage
\section{Responsible Research and Existing Assets}
\label{app:responsible_assets}

\paragraph{Broader impacts.}
This work studies offline model-based reinforcement learning and world modeling in Atari benchmark environments. A potential positive impact is improved compute efficiency for training world models, which may lower the resource barrier for research on sample-efficient decision making. The main risks are indirect: more capable model-based RL systems could eventually be used in domains where inaccurate rollouts, distribution shift, or unsafe planning decisions may cause harm. Our experiments are limited to benchmark video games and do not involve deployment, human-subject data, personal data, or safety-critical decision making. We therefore view the immediate societal risk of this work as limited, while noting that future applications of model-based RL in real-world control should require domain-specific safety evaluation, uncertainty estimation, and deployment safeguards.

\paragraph{Safeguards.}
The paper does not release a high-risk foundation model, image/text generator, scraped dataset, personal dataset, or deployed decision-making system. The code release is intended for reproducing Atari benchmark experiments. The datasets used are public benchmark datasets, and the trained models operate in Atari-like simulated environments. As a result, we do not identify a need for special access restrictions beyond standard research-use documentation and citation of the underlying assets.

\paragraph{Existing assets and licenses.}
We use the public \texttt{rlu-atari} version of the Atari replay data from RL Unplugged, which follows the DQN Replay data distribution used in prior offline Atari work. We cite the original DQN Replay Dataset, RL Unplugged, JOWA, and other baseline methods used for comparison. We also reuse the pretrained tokenizer interface from JOWA for a controlled comparison within the same tokenized world-action framework. Our supplementary code release will include instructions for obtaining the public datasets and pretrained components from their original sources, and users should follow the corresponding licenses and terms of use of those upstream assets. We do not introduce a new dataset.



\end{document}